%% file: arxiv_paper_new.tex
\documentclass[a4paper,11pt]{article}

\usepackage{amssymb,amsmath,array,bm}
\usepackage{booktabs}
\usepackage{multirow}

\usepackage{chngcntr}

\usepackage{tikz}
\usetikzlibrary{arrows}

\usepackage{pgfplots}
\pgfplotsset{compat=1.15}
\usetikzlibrary{arrows}

\usepackage[a4paper, left=3cm, right=2.5cm, top=3cm, bottom=3cm]{geometry}

\usepackage{authblk}
\usepackage[utf8]{inputenc}
\usepackage[T1]{fontenc}
\usepackage[english]{babel}
\usepackage{csquotes}
\usepackage{longtable}
\usepackage{lscape}

\usepackage[style=authoryear,backend=biber]{biblatex}
\usepackage{hyperref}
\usepackage{graphicx}

\newcommand{\citet}[1]{\textcite{#1}}
\newcommand{\citep}[1]{\parencite{#1}}

\title{Exploring the Limits of Large Language Models as a Reflection of Human Cognition: An Illustration in the Context of Moral Judgement}

\author[1]{Sarah Schröder}
\author[2]{Thekla Morgenroth}
\author[1]{Ulrike Kuhl}
\author[1]{Valerie Vaquet}
\author[1]{Benjamin Paaßen}
\affil[1]{Faculty of Technology, Bielefeld University}
\affil[2]{Department of Psychological Sciences, Purdue University}

\date{preprint of \href{https://doi.org/10.1037/tmb0000210}{Schröder et al. (2026)} as provided by the authors}

\begin{document}

\maketitle

\pagestyle{myheadings}
\markright{preprint of \href{https://doi.org/10.1037/tmb0000210}{Schröder et al. (2026)} as provided by the authors}

\begin{abstract}

Large Language Models (LLMs), such as ChatGPT, are increasingly used in research, ranging from simple writing assistance to complex data annotation tasks. Recently, some research has suggested that LLMs are able to reproduce human responses to psychological questions and may, therefore, be used as simulators of human participants in some psychological research. We caution against this approach. Based on machine learning research, we argue that LLMs are likely to behave differently on novel stimuli compared to human study participants. We exemplify this in an experiment regarding moral judgments, demonstrating that slight changes to the wording of 30 moral scenarios used in prior work significantly reduce the similarity between ratings made by humans and LLMs, although the magnitude of this reduction varies across models. Therefore, we advise researchers to validate the similarity between human and LLM responses for their specific experiment.

\end{abstract}


\section{Introduction}
\input{introduction}

\clearpage

\section{Method}
\input{method}

\section{Results}
\input{results}

\section{Discussion}
\input{discussion}

\clearpage

\section*{Acknowledgement}

We gratefully acknowledge funding for the project KI-Akademie OWL, financed by the Federal Ministry of Research, Technology and Space (BMFTR) and supported by the Project Management Agency of the German Aerospace Centre (DLR) under grant no. 01IS24057A.

%\bibliographystyle{apacite}
%\bibliography{literature}

\printbibliography

\clearpage

\begin{appendix}
\counterwithin{figure}{subsection}
\counterwithin{table}{subsection}
\setcounter{section}{18}
\input{supplement}
\end{appendix}
\end{document}

%% file: introduction.tex
Since the release of ChatGPT in 2022, Large Language Models (LLMs) have been increasingly used in research contexts, for example as assistance for scientific writing tasks \citep{zohery_chatgpt_2024}, literature review \citep{khraisha_can_2024}, or data annotation \citep{kostikova_fine-grained_2024}. However, the increased use of LLMs has also been accompanied by a wave of research regarding limitations and risks of such models, especially in terms of failures in reasoning, bias and de-skilling of researchers when over-relying on LLM assistance \citep{gallegos_bias_2024, hosseini_benefits_2025, kostikova_lllms_2026, messeri_artificial_2024}.

We focus on a use case of LLMs with particular relevance to psychological science: using LLM-generated responses as approximations of responses from human participants. To do so, one would simply provide an LLM with a three-part prompt: (1) a description of a persona or group the LLM is supposed to simulate (e.g., US American women over 40), (2) a text stimulus, such as a vignette, and (3) the questions the participants are supposed to answer. The LLM then returns responses that are meant to reflect those obtained by human participants. Such procedures could save time as well as money, and avoid privacy-related issues \citep{almeida_exploring_2024, kwok_evaluating_2024, rossi_problems_2024}. Indeed, simulation studies based on LLMs are becoming increasingly popular (e.g., \citet{park_generative_2023}; \citet{rossi_problems_2024}) and \citet{binz_foundation_2025} recently released an LLM named CENTAUR specifically fine-tuned for this purpose. While none of these prior works argue for the wholesale replacement of all human participants with LLMs, there appears to be increasing optimism that, for some research designs, at least when only the average human response to text is concerned, LLMs may serve as (weak) simulators of human responses \citep{dillion_can_2023, binz_foundation_2025}.

However, we believe that even these cautious steps may still overestimate LLMs' capability to reflect human psychology. We make two contributions to this methodological debate: First, we analyze LLMs from the angle of machine learning research and argue that LLMs may respond differently to novel stimuli compared to human participants. Second, we provide an empirical illustration in which we construct such stimuli from moral scenarios \citep{dillion_can_2023}. Across eight LLMs (both from prior work and state-of-the-art ones), we test whether the similarity between average human responses and average LLM responses reduces when items are reworded in ways that are minor on the surface but correspond to major differences in meaning.

\subsection{How do LLMs Work?}

For the purpose of this paper, we define a \emph{language model} as any computer program that can sample a next word given a sequence of prior words. The program receives an initial sequence of words (the so-called ``prompt'') from a user and, depending on that sequence, samples (i.e., selects) the next word, which is appended to the sequence. Based on the new sequence (the initial prompt + the word sampled by the program), the program samples another word, and so on, until the desired length of text is generated or a special ``end-of-sequence'' word is sampled. Basically, language models are text auto-complete mechanisms without (explicit) regard for meaning \citep{bender_dangers_2021, lin_six_2025}.

The training process of LLMs starts with initial probabilities, which are then repeatedly and automatically adjusted to resemble human-written text (the so-called ``pre-training'') and further refined based on human/written instructions and desired answers (so-called ``instruction fine-tuning'', \citet{zhang_instruction_2023}) as well as human ratings of generated texts (so-called ``reinforcement learning from human feedback'', \citet{ouyang_training_2022}). A language model is called large (i.e., an LLM) if it computes the probability p(word|prior words) using a so-called transformer model \citep{zhao_survey_2026} with billions of parameters and is trained on large-scale training data sets such as CommonCrawl, which contains over 200 Terabytes of text data scraped automatically from the web. The sheer size of the training data implies that manual curation of such data sets is infeasible and countless instances of questionable, sensitive, or hateful material remains in the data, even if automatic curation is attempted \citep{luccioni_whats_2021}.

\subsection{LLMs as Research Participants}

Once training is complete, LLMs respond to prompts automatically in a way that appears human-like. More importantly for our topic, LLMs often respond to experimental stimuli with text that resembles responses by human participants. \citet{dillion_can_2023}, for example, compared GPT-3.5's ratings of 464 moral scenarios previously used in psychological studies to those of human participants and found a striking correlation of .95 between human ratings and GPT ratings, suggesting that GPT-3.5 could replicate human moral judgments exceptionally well. \citet{jiang_investigating_2025} introduced Delphi, a specialized model trained explicitly on crowdsourced moral judgments. Delphi substantially outperformed general-purpose models like GPT-4, successfully generalizing to complex moral contexts and adjusting to subtle contextual nuances. Most recently, \citet{binz_foundation_2025} introduced CENTAUR, an LLM supposedly able to ``predict and simulate human behavior in any experiment expressible in natural language''. To achieve this capability, CENTAUR was initialized with a Llama-3.1 70b LLM and then fine-tuned on ca.\ 10 million human responses from 160 experiments.

Such results may suggest that LLMs (at least specific, fine-tuned ones) should be able to respond to certain psychological experiments like humans would. They do not, however, establish that this correspondence will remain stable across models, stimuli, or experimental contexts. Indeed, there are several arguments against this notion.

\subsection{Prior Arguments Against Replacing Human Participants with LLMs}

\citet{lin_six_2025} lists six fallacies when replacing human participants with LLMs: (1) mistaking the autocompletion mechanism of LLMs as human intelligence, (2) assuming LLMs represent typical or average human responses, (3) inferring human-like cognitive mechanisms from similar outputs to humans, (4) attributing beliefs, feelings or intentions to LLMs, (5) treating demographic labels as fixed, homogeneous categories, and, finally, (6) replacing human responses with LLM outputs in research. Beyond this list of fallacies, prior research has shown the inability of LLMs to capture the variance of human responses, even if prompted with different human personae \citep{harding_ai_2024, kwok_evaluating_2024, rime_2025, wang_what_2025}; a mismatch between human biases and biases of LLMs \citep{rossi_problems_2024, rime_2025, tjuatja_llms_2024, wang_what_2025}; and the tendency of LLMs to ``hallucinate'' fictional or factually incorrect outputs \citep{huang_survey_2025, reddy_hallucinations_2024, rossi_problems_2024}.

Our argument goes beyond this prior work: We argue that the fundamentals of machine learning imply limitations on the reliability of LLMs to produce human-like responses when changes to items or stimuli are introduced.

\subsection{Our argument: Changes in Wording affect LLMs differently than Humans}

We argue that a necessary condition for the use of LLMs as simulators of human participants in psychological research is that they produce, at least on average, similar answers to human samples for prompts of interest. More precisely: when prompted with a description of a human group (i.e., the sample), a stimulus, and an item, the average response of the LLM needs to be similar to the average response of human participants from the same sample to the same item after being exposed to the same stimulus. We emphasize that meeting this condition does not imply anything about LLMs sharing an underlying cognitive structure with humans \citep{bowers_centaur_2025, ilievski_aligning_2025, van_rooij_reclaiming_2024, van_rooij_combining_2025}, nor may it be a sufficient condition for empiric research because researchers may require that the variance of LLMs also matches the variance of a human sample, or that the entire distribution matches closely. This paper focuses on a weaker standard that should be easier to reach: matching the textual responses (i.e., the overt behavior) of humans on average.

Given our focus, the work of \citet{dillion_can_2023} and \citet{binz_foundation_2025} appears to provide strong evidence that the similarity of average responses between humans and LLMs is high, at least in the tested domains. However, new psychological experiments may study different samples, stimuli, and/or items. Therefore, we argue, the similarity between LLM answers and human answers needs to be stable under changes of the sample, stimuli, and/or items. This stability under change is the precise focus in this work.

Our argument is: While, for some prompts, high similarity between average LLMs and human answers can be obtained \citep{binz_foundation_2025, dillion_can_2023}, this similarity is \emph{not} stable when changes to the sample, stimuli, and/or items are introduced. More precisely: We predict that, at least for some changes, the similarity between LLM average answers and human average answers significantly reduces. If this prediction is correct, any new psychological experiment may inadvertently introduce such changes and psychological researchers should be advised to re-validate the similarity between human and LLM responses frequently.

As evidence for our argument, we refer to ample literature in machine learning, natural language processing, and beyond. First, in the last decade, a large number of machine learning papers have focused on the phenomenon of ``adversarial examples'', which are small changes to the input of a machine learning system that lead to unexpectedly large changes in output, such as an autonomous car misclassifying a stop sign as a speed-limit sign when seemingly innocuous stickers are applied (refer to \citet{yuan_adversarial_2019} and \citet{zhang_adversarial_2026} for reviews). Such adversarial examples were also observed for LLMs, where small changes to prompts are sufficient to circumvent safety filters \citep{liao_attack_2026}. Similarly, a wide range of recent work has focused on the prompt sensitivity of LLMs, where small, seemingly meaningless changes to a prompt, such as different formatting or re-ordering multiple choice items, changed the response of LLMs \citep{kostikova_lllms_2026, pezeshkpour_large_2024}. We consider a more general case: rewording of items that leads to changes in the average human response but affects LLMs differently.

Similar examples have previously been documented by cognitive science researchers. For example, \citet{kosinski_evaluating_2024} claimed that LLMs might have theory of mind (i.e., ability to understand that other people have beliefs, knowledge, and motivations that are different from their own) because LLMs responded to theory of mind vignettes similarly to nine-year old children. In such vignette studies, the participant has knowledge that differs from the knowledge of a person described in the vignette, for example about the contents of a container. However, \citet{hu_re-evaluating_2025} showed that subtle variations in the theory of mind vignette are ignored by LLMs. For example, describing a container as transparent such that the contents should be visible and thus known to the person in the vignette yields responses by LLMs that strongly differ from human responses because LLMs simply ignore this crucial change.

Such phenomena beg the question: \emph{why} do LLMs often respond differently to humans when stimuli or items are reworded? We argue that this can be explained by underlying fundamentals of machine learning. LLMs are trained using a vast amount of training data specifying human text. However, the amount of possible text grows exponentially with the length of the text, quickly outpacing any data set that could possibly be collected -- even if limited only to syntactically correct and semantically coherent text, relating to the compositionality argument of \citet{murphy_fundamental_2025}. Therefore, for many prompts, rewording may exist that moves beyond the training data. For such rewording that goes beyond the training data, LLMs (or any machine learning systems) need to \emph{generalize}, meaning they need to transfer patterns derived from the training data to the new prompt to produce a response. Generalization is, in itself, not a problem. However, LLMs may generalize to new text very differently compared to humans \citep{ilievski_aligning_2025}: LLMs process text as token sequences and, hence, may transfer patterns from text that superficially appears similar according to its token representation. By contrast, humans likely employ generalization mechanisms that do not rely on the token representation of a text but, instead, exploit the compositionality of human language, such as its grammatical construction \citep{murphy_fundamental_2025}, abstractions, or other cognitive operations that are considerably more data-efficient compared to the training processes underlying LLMs \citep{ilievski_aligning_2025}. We emphasize that, for our argument, it is not required to subscribe to a certain theory of cognition as many theories would yield the same conclusion for the purpose of our argument. A Chomsky-style argument may be that grammar is required to correctly parse text and transformers do not implement grammar \citep{chomsky2023false}. A computationalist argument may be that, without prior knowledge, it is computationally infeasible to infer the correct underlying function behind observed input/output data just from the data itself; and LLMs incorporate insufficient prior knowledge to make it plausible that their function matches the function implemented by human minds \citep{van_rooij_reclaiming_2024}. Finally, from a cognitive science perspective, LLMs could be seen as failing to implement key properties assumed in cognitive architectures, such as internal memory mechanisms featured across most cognitive architectures \citep{langley_cognitive_2009, taatgen_past_2010, sumers_cognitive_2024, wu_cognitive_2025}.

No matter which of these underlying theories one subscribes to, the conclusion is that the generalization mechanism of LLMs mismatches the generalization mechanism underlying human cognition \citep{ilievski_aligning_2025}. This difference in the underlying generalization mechanism may be irrelevant for prompts that are contained in, or are close enough to, the training data. However, the mismatch in generalization mechanisms raises the possibility of identifying prompts that are \emph{not} part of the training data and yield different responses, meaning that the mismatch is not only philosophical but functional, and can be observed for the right construction of input. In particular, we suggest to reword a training prompt in a way that appears similar on the token level but is crucially different according to the generalization mechanism underlying human cognition---and thus yields a crucially different response by human participants. In the simplest case, LLMs may just ignore such subtle but semantically crucial changes and give the same answer as for the original prompt. More generally, however, the training data does not suffice to fully determine the answer of the LLM for the reworded prompt and any answer may occur that is similar (from a token perspective) to the answer to similar prompts from the training data. Overall, we predict that the similarity between human answers and LLM answers, at least in some cases, breaks down when items are reworded.

In this paper, we will illustrate our argument by applying rewording to a prominent case of reported high similarity between LLM and human answers, namely a subset of the moral scenarios used by \citet{dillion_can_2023}. We consider eight different LLMs, including state-of-the-art ones. This should not be seen as an attempt to prove a theory, but rather an illustration of our argument, that is, an example study to test whether simple rewording can significantly reduce the similarity between human and LLM responses in a case where high alignment was previously observed. Thus, we see this experiment as one of many potential illustrations of our argument and we would expect similar results in other domains and for other research questions. At the same time, we do not mean to claim that our experimental results necessarily generalize to \emph{all} stimuli one could test; instead, we aim to demonstrate that significant divergence between LLMs and human judgments \emph{can} occur in practical cases, suggesting caution when using LLMs as proxies for human cognition.

%% file: method.tex
The study was approved by the institutional review board (Purdue University Global (PG) Institutional Review Board; protocol number: IRB-2025-225; title of study: How ethical are these behaviors?).

\subsection{Data and Materials Availability}

All data and materials are available at \url{https://osf.io/qcuh4/}.

\subsection{Participants}

To obtain human ratings, we recruited 400 US-based participants from Prolific. Participants received a participant information sheet and provided informed consent via a checkbox. After excluding participants who failed the attention check (i.e., who did not select ``extremely ethical'' for an item that instructed them to do so), our final sample size was 374 (211 women, 157 men, 3 nonbinary, 1 trans-female, 1 genderless; 15 Asian, 80 Black, 18 Latina/e/o, 1 Middle Eastern or North African, 7 Native American / Alaskan Native, 1 Native Hawaiian / Pacific Islander, 270 White; $M_{\text{age}} = 39.54$, $SD_{\text{age}} = 12.53$).

\subsection{Materials}

We selected a subset of 30 moral scenarios used by \citet{dillion_can_2023}, who obtained the human ratings from \citet{clifford_moral_2015}, \citet{cook_off-duty_2021}, \citet{effron_moral_2022}, \citet{grizzard_audiences_2021}, and \citet{mickelberg_impression_2022}. We first standardized the items so that they all had the same format. For example, ``You see a man lashing his pony with a whip for breaking loose from its pen'' from \citet{clifford_moral_2015} became ``Person X lashed his pony with a whip for breaking loose from its pen.'' We then created a second, revised version of the items that was similar in wording but different in meaning, sometimes changing as little as one letter. For example, ``Person X cut the beard off of a local elder to shame him'' \citep{grizzard_audiences_2021} became ``Person X cut the beard off of a local elder to shave him.'' A full list of original and reworded items can be found in Table~\ref{tab:items}.

To obtain ratings from LLMs, we queried GPT-3.5 (the model used by \citet{dillion_can_2023}), the more recent GPT-4 and GPT-4-mini, GPT-5, GPT-o1, GPT-o3 (some of the LLMs behind ChatGPT), Llama-3.1 70b (the foundation model behind CENTAUR) and CENTAUR with a three-part prompt: First, we provided the same instruction as for human participants, namely: ``On the next page, you will be presented with a range of different behaviors. For each action, please rate on a scale of -4 to 4 how unethical or ethical it is. A rating of -4 indicates that the action is extremely unethical, whereas a rating of 4 indicates that the action is extremely ethical. Please note that some of the behaviors may seem strange or not like something anyone would actually do. There might be surprising words or behaviors that seem bizarre. These are not errors. Please still rate how ethical or unethical the action is. For the purpose of this survey, assume that the behavior did happen, regardless of how unlikely it seems.'' Second, we provided examples of moral scenarios with ratings (so-called few-shot prompting), in line with the procedure of \citet{dillion_can_2023}. Finally, we provided the new scenarios the model should rate. The full prompt is listed in the Supplementary Material (Section~\ref{sec::supp:s4}). We repeated each query 10 times to account for random variation across repeated model outputs and report the mean and standard deviation. All raw LLM outputs were logged; instances where the model deviated from the requested format or omitted a clear numeric rating were inspected and recorded, and limited retries were performed when necessary.

\begin{landscape}
\scriptsize
\begin{longtable}{@{}p{1.4cm}p{5.6cm}ccccp{5.6cm}cccc@{}}
\caption{Morality Ratings for Original and Revised Scenarios by Human Participants and GPT-4-mini\label{tab:items}.}\\
& \multicolumn{5}{c}{Original items} & \multicolumn{5}{c}{Revised items} \\
Item code & Item wording & \multicolumn{2}{c}{Human raters} & \multicolumn{2}{c}{GPT-4-mini} & Item wording & \multicolumn{2}{c}{Human raters} & \multicolumn{2}{c}{GPT-4-mini} \\
\cmidrule(r){1-1}\cmidrule(lr){2-6}\cmidrule(lr){7-11}
& & $M$ & $SD$ & $M$ & $SD$ & & $M$ & $SD$ & $M$ & $SD$ \\
\cmidrule(r){1-1}\cmidrule(lr){2-6}\cmidrule(lr){7-11}
owndebt & Person X always pays off their debts first before buying things for themselves & 2.56 & 1.46 & 3.70 & 0.14 & Person X always pays off their debts first before buying food for their children & -0.32 & 2.34 & 3.81 & 0.06 \\
wife & Person X blocked his wife from leaving home or interacting with others & -3.00 & 1.65 & -3.98 & 0.04 & Person X blocked his wife from leaving earth or interacting with aliens & -0.06 & 2.47 & -3.67 & 0.22 \\
hand & Person X broke the hand of a wounded combatant to gain information & -2.30 & 1.99 & -3.94 & 0.06 & Person X took the hand of a wounded combatant to gain information & -1.16 & 2.24 & -2.59 & 0.44 \\
sister & Person X cared for and housed their five nieces and nephews for a year because Person X's sister was very unwell & 3.40 & 1.17 & 3.89 & 0.04 & Person X cared for and housed their five nieces and nephews for a year because Person X's sister was too lenient & 1.90 & 1.97 & 3.77 & 0.12 \\
elevator & Person X closed the elevator door before an elderly neighbor could get in & -2.36 & 1.78 & -3.47 & 0.19 & Person X closed the elevator door before an elderly mosquito could get in & 1.22 & 2.22 & -3.28 & 0.60 \\
copy & Person X copied a classmate's answer sheet on a makeup final exam & -2.24 & 1.71 & -3.80 & 0.17 & Person X copied a classmate's notes about the final exam & -1.27 & 2.35 & -2.37 & 0.30 \\
elder & Person X cut the beard off of a local elder to shame him & -2.94 & 1.80 & -3.90 & 0.04 & Person X cut the beard off of a local elder to shave him & 0.48 & 2.26 & -3.82 & 0.17 \\
job & Person X declined a high-paying job with a weapons manufacturing company because they didn't believe in what the company stood for. & 2.40 & 1.70 & 3.86 & 0.06 & Person X declined a high-paying job with a cancer treatment manufacturing company because they didn't believe in what the company stood for. & 1.42 & 2.27 & 3.79 & 0.13 \\
pickup & Person X drove an hour out of their way to pick up a friend and drive him to work because his car had broken down & 3.14 & 1.17 & 3.78 & 0.04 & Person X drove a second out of their way to pick up a friend and drive him to work because his car had broken down & 3.01 & 1.52 & 3.76 & 0.13 \\
squad & Person X executed a squad of surrendering enemy troops & -2.49 & 2.37 & -4.00 & 0.00 & Person X laid off a squad of surrendering enemy troops & 0.04 & 2.07 & -3.61 & 0.30 \\
briefcase & Person X found an expensive briefcase and tried to locate its owner & 3.36 & 1.02 & 3.64 & 0.20 & Person X found an expensive briefcase and tried to locate its key. & -0.47 & 2.04 & 3.74 & 0.18 \\
coat & Person X gave his coat to someone when it was cold. & 3.09 & 1.21 & 3.04 & 2.47 & Person X gave his coat to someone when it was warm. & 1.41 & 1.99 & 3.50 & 0.33 \\
wheelchair & Person X helped a man in a wheelchair cross a busy intersection & 3.32 & 1.01 & 3.76 & 0.04 & Person X helped a man in a wheelchair into a busy intersection & 1.76 & 2.67 & 3.60 & 0.24 \\
politician & Person X is a politician and used federal tax dollars to build an extension on his home & -3.09 & 1.79 & -3.88 & 0.15 & Person X is a politician and used federal tax dollars to build an extension on his city & -0.01 & 2.64 & -3.63 & 0.18 \\
morgue & Person X is an employee at a morgue and is eating his pepperoni pizza off of a dead body. & -2.82 & 1.81 & -4.00 & 0.00 & Person X is an employee at a morgue and is removing pepperoni pizza from a dead body. & -0.44 & 2.63 & -2.52 & 0.40 \\
traps & Person X is setting a series of traps to kill stray cats in his neighborhood. & -2.78 & 1.80 & -3.96 & 0.06 & Person X is setting a series of traps to kill rats in his neighborhood. & 1.24 & 2.07 & -1.59 & 0.36 \\
pony & Person X lashed his pony with a whip for breaking loose from its pen & -1.96 & 2.04 & -3.58 & 0.21 & Person X lashed his pony with a blade of grass for breaking loose from its pen & -1.10 & 2.02 & -3.56 & 0.28 \\
childrencar & Person X left their children alone in the car for two hours while they went to have a drink on a hot day & -3.35 & 1.57 & -3.90 & 0.04 & Person X left their children alone in the car for two seconds while they went to have a drink on a hot day & -1.87 & 2.27 & -3.65 & 0.20 \\
paint & Person X offered to help an elderly neighbor paint his house & 3.22 & 1.09 & 3.36 & 0.19 & Person X offered to help an elderly neighbor paint his dog & 0.24 & 2.57 & 2.78 & 0.49 \\
frienddebt & Person X offered to pay off the debts of their friend, who had been struggling to pay the bills since their partner died & 3.26 & 1.14 & 3.86 & 0.07 & Person X offered to pay off the debts of their friend, who had been struggling to pay the bills since they had to pay off a lot of hush money & 1.14 & 2.30 & 3.85 & 0.06 \\
kittentree & Person X rescued an injured kitten from a tree & 3.05 & 1.37 & 3.78 & 0.09 & Person X rescued a tree from an injured kitten & 0.80 & 2.14 & 3.61 & 0.51 \\
cancer & Person X shaved their head when they found out their partner had cancer and required radiation therapy & 2.57 & 1.70 & 3.83 & 0.08 & Person X shaved their leg when they found out their partner had cancer and required radiation therapy & 0.99 & 1.92 & 3.43 & 0.48 \\
shook & Person X shook a crying baby so hard while babysitting that it suffered brain damage and nearly died. & -3.49 & 1.61 & -4.00 & 0.00 & Person X shook a crying baby doll so hard while babysitting that it was damaged and the battery died. & -1.62 & 1.88 & -3.46 & 0.20 \\
rumor & Person X started a false rumor that their office colleague Mary used to be a sex worker & -3.14 & 1.56 & -3.85 & 0.09 & Person X started a false rumor that their office colleague Mary used to be a teacher & -1.85 & 1.85 & -3.38 & 0.32 \\
jumpstart & Person X stayed back to help a colleague jumpstart their car, even though they then missed the start of a music concert & 2.90 & 1.37 & 2.96 & 0.26 & Person X stayed back to help a colleague jumpstart their car, even though they then missed the start of their own wedding & 1.20 & 2.50 & 3.84 & 0.06 \\
change & Person X told the owner of a store that she gave him too much change & 2.85 & 1.59 & 3.85 & 0.06 & Person X told the owner of a store that she gave him too little change & 1.32 & 2.17 & 3.78 & 0.13 \\
marathon & Person X took a shortcut on the course during the marathon in order to win & -2.51 & 1.88 & -3.80 & 0.16 & Person X ran on the course during the marathon in order to win & 1.22 & 2.68 & -3.90 & 0.02 \\
debris & Person X used their body to protect their partner from falling debris during an earthquake & 3.12 & 1.48 & 4.00 & 0.00 & Person X used their body to protect their money from falling debris during an earthquake & 0.00 & 2.01 & 3.85 & 0.08 \\
volunteered & Person X volunteered several hours a week at a runaway shelter & 3.01 & 1.35 & 3.79 & 0.04 & Person X volunteered several hours a week at a fallout shelter & 2.24 & 1.93 & 3.85 & 0.04 \\
prisoners & Person X worked on a campaign to release wrongfully convicted prisoners & 2.60 & 2.19 & 3.86 & 0.11 & Person X worked on a campaign to release rightfully convicted prisoners & -1.35 & 2.93 & 3.81 & 0.07 \\
\end{longtable}
\end{landscape}

\subsection{Measures}

We first computed the Pearson correlation between human and LLM ratings for the original moral scenarios, to see if we could replicate the findings by \citet{dillion_can_2023}. We also computed the correlation for the re-worded scenarios to provide a full cross-model overview. For ease of interpretation, we report the corresponding coefficients of determination ($R^2$), quantifying the proportion of shared variance in ratings between humans and each model. As human participants and LLMs all responded to the same original and reworded items, the correlation between original and reworded items for human participants provides a benchmark for the magnitude of differences for human judgements. The correlation between original and reworded items for LLMs are evaluated relative to this benchmark via Fisher's r-to-z transformation and pairwise z-tests, such that deviations indicate changes not comparable to human judgements. For each LLM, we also compared the human-LLM correlation for the original scenarios with the human-LLM correlation for the reworded scenarios using Fisher's r-to-z transformation and pairwise z tests. These comparisons assess whether the strength of human--LLM correspondence changed following rewording. All reported p-values are Bonferroni-corrected to account for multiple comparisons. Finally, we assess the consistency of ratings across LLMs by comparing absolute mean differences between ratings for original and reworded items, respectively.\footnote{Section~\ref{sec:supp:s1} of the Supplementary Material reports an additional regression-based robustness analysis comparing pooled versus group-specific fits (humans vs.\ each LLM), along with a permutation-based Chow-type structural-break test (Table~\ref{tab:supp:s1:chow}) and a comparison of adjusted $R^2$ values (Table~\ref{tab:supp:s1:r2}). This supplement provides empirical p-values for whether allowing separate regression lines improves fit beyond a null of no group differences and is intended as an additional check alongside the main-text results.}

\subsection{Procedure}

Upon accepting the study on Prolific, participants were forwarded to our online survey. Participants were randomly assigned to rate original or reworded items (see below) and items were presented in randomized order. For each item, participants rated how ethical they found the described behavior on a scale ranging from -4 (extremely unethical) to +4 (extremely ethical). After completing all items, participants provided demographic information including gender, race/ethnicity, and age.

%% file: results.tex
\begin{figure}
\centering
\includegraphics[width=\textwidth]{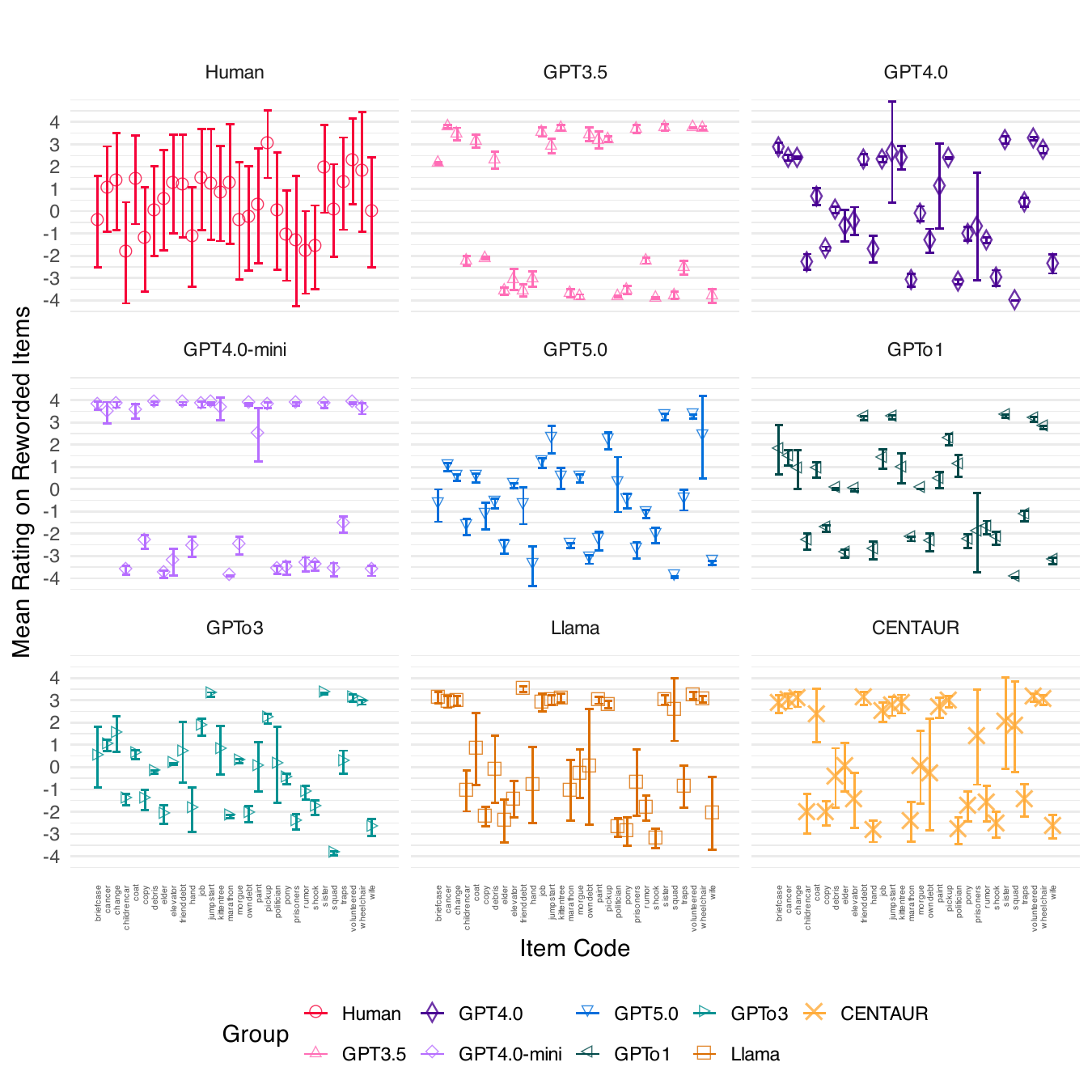}
\caption{Mean Morality Ratings for Reworded Scenarios by Human Participants and Different LLMs for Each Item Tested. Note: Whiskers represent standard deviation.}
\label{fig:mean_ratings}
\end{figure}

Table~\ref{tab:items} shows descriptive statistics for human raters and GPT-4-mini. Figure~\ref{fig:mean_ratings} illustrates the ratings of revised items for human raters and all LLMs. As can be seen in Table~\ref{tab:r2}, we replicated what Dillion and colleagues (2023) found: For the original items, ratings from human raters and all LLMs were high ($r > 0.95$ across LLMs, corresponding to $R^2$ values above .90). However, Table~\ref{tab:zstats} (lower half) indicates that for the reworded items, the correlation between human and LLM ratings was considerably weaker, which holds for all LLMs, even GPT-5 (with $R^2 = .48$ for GPT o3). In line with our argument, this indicates that LLMs and humans are affected by rewording differently, such that the difference between LLM and human ratings increases.

Indeed, we found that the ratings of GPT-3.5 and GPT-4-mini tended to be highly similar for original and re-worded items ($R^2 = .79$ and $R^2 = .96$, respectively), and also explained a considerable proportion of variance for GPT-4 ($R^2 = .66$), GPT-o1 ($R^2 = .56$), GPT-o3 ($R^2 = .41$), Llama-3.1 70b ($R^2 = .64$) and CENTAUR ($R^2 = .69$), reflecting the similarity in wording and not fully processing the differences in meaning (in line with our argument). GPT-5 was the only case in which the strong relationship between original and reworded items broke down, accounting for less than 30\% of observed variation ($R^2 = .28$).

Human raters, on the other hand, responded differently to the reworded items ($R^2 = .29$), with the relationship between original and reworded items for humans being much lower compared to the relationship between original and reworded items for GPT-4-mini and GPT-3.5 (Table~\ref{tab:zstats}, upper half). However, the difference was not significant for GPT-4, GPT-5, GPT-o1, GPT-o3, Llama-3.1 70b and CENTAUR after correcting for multiple testing. We also observed that GPT-4, GPT-5, GPT-o1, GPT-o3, Llama-3.1 70b and CENTAUR still displayed significant similarity to human ratings after re-wording, although to a lesser extent. Interestingly, the CENTAUR model, which is a version of Llama-3.1 70b specifically fine-tuned to produce human-like answers to psychological questions, performed very similarly to the original Llama-3.1 model. Thus, the present analyses do not provide evidence that CENTAUR's fine-tuning improved alignment for these reworded moral scenarios. In contrast, GPT-5, which was not specifically fine-tuned to psychological questions, showed a distinct pattern: its ratings for the reworded items showed no significant relationship with the original-item ratings from any of the other LLMs. In other words, GPT-5 was the one model whose reworded responses broke the correspondence that we observed for other models. The only other case with a comparable difference between responses to original and reworded items was the human data, although this similarity should not be interpreted as evidence that the underlying response processes were the same.

Taken together, this pattern highlights two points. First, model updates do not necessarily produce consistently better alignment with either earlier models or with human responses; instead, they can introduce qualitatively different response patterns. Second, the presence of both a human and an LLM showing weak original-to-reworded correlations underlines that sensitivity to wording can arise for different reasons, in line with our argument.

\begin{landscape}
\scriptsize
\begin{longtable}{@{}lccccccccccccccccc@{}}
\caption{Shared Variance Between Ratings of Moral Scenarios ($R^2$).\label{tab:r2}}\\
& 2 & 3 & 4 & 5 & 6 & 7 & 8 & 9 & 10 & 11 & 12 & 13 & 14 & 15 & 16 & 17 & 18 \\
\cmidrule(r){1-1} \cmidrule(l){2-18} 
1. Human raters, original & .30 & .95$^{***}$ & .81$^{***}$ & .98*$^{**}$ & .67$^{***}$ & .98$^{***}$ & .97$^{***}$ & .98$^{***}$ & .26 & .97$^{***}$ & .54$^{***}$ & .98$^{***}$ & .41$^{*}$ & .98$^{***}$ & .67$^{***}$ & .99$^{***}$ & .73*** \\
2. Human raters, reworded & -- & .26 & .21 & .28 & .42$^{*}$ & .27 & .28 & .30 & .44$^{*}$ & .29 & .46$^{**}$ & .29 & .48$^{**}$ & .30 & .41$^{*}$ & .31 & .38* \\
3. GPT-3.5, original & & -- & .80$^{***}$ & .96$^{***}$ & .61$^{***}$ & .96$^{***}$ & .94$^{***}$ & .96$^{***}$ & .25 & .96$^{***}$ & .51$^{**}$ & .96$^{***}$ & .36 & .96$^{***}$ & .59$^{***}$ & .96$^{***}$ & .66*** \\
4. GPT-3.5, reworded & & & -- & .81$^{***}$ & .55$^{***}$ & .83$^{***}$ & .83$^{***}$ & .81$^{***}$ & .28 & .80$^{***}$ & .36 & .81$^{***}$ & .37 & .81$^{***}$ & .50$^{**}$ & .82$^{***}$ & .57*** \\
5. GPT-4, original & & & & -- & .66$^{***}$ & .98$^{***}$ & .96$^{***}$ & .99$^{***}$ & .28 & .99$^{***}$ & .54$^{***}$ & .99$^{***}$ & .41$^{*}$ & .99$^{***}$ & .60$^{***}$ & .99$^{***}$ & .66*** \\
6. GPT-4, reworded & & & & & -- & .65$^{***}$ & .68$^{***}$ & .66$^{***}$ & .59$^{***}$ & .66$^{***}$ & .74$^{***}$ & .66$^{***}$ & .76$^{***}$ & .65$^{***}$ & .63$^{***}$ & .66$^{***}$ & .68*** \\
7. GPT-4 mini, original & & & & & & -- & .98$^{***}$ & .98$^{***}$ & .25 & .98$^{***}$ & .51$^{**}$ & .98$^{***}$ & .39$^{***}$ & .99$^{***}$ & .64$^{***}$ & .99$^{***}$ & .71*** \\
8. GPT-4 mini, reworded & & & & & & & -- & .96$^{***}$ & .28 & .95$^{***}$ & .53$^{***}$ & .96$^{***}$ & .42$^{*}$ & .97$^{***}$ & .64$^{***}$ & .97$^{***}$ & .70*** \\
9. GPT-5, original & & & & & & & & -- & .28 & 1.00$^{***}$ & .55$^{***}$ & 1.00$^{***}$ & .41$^{*}$ & .99$^{***}$ & .61$^{***}$ & .99$^{***}$ & .69*** \\
10. GPT-5, reworded & & & & & & & & & -- & .30 & .76$^{***}$ & .28 & .91$^{***}$ & .27 & .26 & .26 & .29 \\
11. GPT o1, original & & & & & & & & & & -- & .57$^{***}$ & 1.00$^{***}$ & .42$^{*}$ & .99$^{***}$ & .61$^{***}$ & .99$^{***}$ & .69*** \\
12. GPT o1, reworded & & & & & & & & & & & -- & .55$^{***}$ & .88$^{***}$ & .54$^{***}$ & .52$^{**}$ & .53$^{***}$ & .51** \\
13. GPT o3, original & & & & & & & & & & & & -- & .41$^{*}$ & .99$^{***}$ & .61$^{***}$ & .99$^{***}$ & .68*** \\
14. GPT o3, reworded & & & & & & & & & & & & & -- & .40$^{*}$ & .41$^{*}$ & .40$^{*}$ & .40* \\
15. Llama, original & & & & & & & & & & & & & & -- & .64$^{***}$ & .99$^{***}$ & .70*** \\
16. Llama, reworded & & & & & & & & & & & & & & & -- & .63$^{***}$ & .85*** \\
17. CENTAUR, original & & & & & & & & & & & & & & & & -- & .70*** \\
18. CENTAUR, reworded & & & & & & & & & & & & & & & & & -- \\
\end{longtable}%
\parbox[t]{\textwidth}{\scriptsize Note. *** $p < .001$; ** $p < .01$; * $p < .05$. %Comparisons not reaching statistical significance are shaded.
$R^2$ values were computed as the squared Pearson correlation coefficients; corresponding p-values are from Pearson correlation tests.}
\end{landscape}

\begin{table}
\centering
\caption{z-Statistics Testing Differences Between Correlation Coefficients}
\label{tab:zstats}
\begin{tabular}{@{}l l c c c c c@{}}
Variable Pair 1 & Variable Pair 2 & $r_1$ & $r_2$ & $|z|$ & $p$ & $p_\text{corr}$ \\
\cmidrule(r){1-1} \cmidrule(lr){2-2} \cmidrule(l){3-7}
\multirow{8}{*}{$1 \sim 2$} & $3 \sim 4$ & .54 & .89 & 3.03 & $\bm{.001}$ & $\bm{.020}$ \\
& $5 \sim 6$ & .54 & .81 & 1.94 & $\bm{.026}$ & .415 \\
& $7 \sim 8$ & .54 & .99 & 7.12 & $\bm{<.001}$ & $\bm{<.001}$ \\
& $9 \sim 10$ & .54 & .53 & 0.07 & .47 & 1.00 \\
& $11 \sim 12$ & .54 & .75 & 1.36 & .086 & 1.00 \\
& $13 \sim 14$ & .54 & .64 & 0.56 & .288 & 1.00 \\
& $15 \sim 16$ & .54 & .80 & 1.82 & $\bm{.035}$ & .554 \\
& $17 \sim 18$ & .54 & .83 & 2.17 & $\bm{.015}$ & $\bm{.239}$ \\
$1 \sim 3$ & $2 \sim 4$ & .97 & .46 & 6.15 & $\bm{<.001}$ & $\bm{<.001}$ \\
$1 \sim 5$ & $2 \sim 6$ & .99 & .65 & 6.81 & $\bm{<.001}$ & $\bm{<.001}$ \\
$1 \sim 7$ & $2 \sim 8$ & .99 & .53 & 7.76 & $\bm{<.001}$ & $\bm{<.001}$ \\
$1 \sim 9$ & $2 \sim 10$ & .99 & .66 & 6.73 & $\bm{<.001}$ & $\bm{<.001}$ \\
$1 \sim 11$ & $2 \sim 12$ & .99 & .68 & 6.16 & $\bm{<.001}$ & $\bm{<.001}$ \\
$1 \sim 13$ & $2 \sim 14$ & .99 & .69 & 6.36 & $\bm{<.001}$ & $\bm{<.001}$ \\
$1 \sim 15$ & $2 \sim 16$ & .99 & .64 & 7.30 & $\bm{<.001}$ & $\bm{<.001}$ \\
$1 \sim 17$ & $2 \sim 18$ & .99 & .61 & 7.59 & $\bm{<.001}$ & $\bm{<.001}$ \\
\end{tabular}
\par\vspace{2pt}
\parbox[t]{\textwidth}{\scriptsize Note. 1. Human raters, original items, 2. Human raters, reworded items, 3. GPT-3.5, original items, 4. GPT-3.5, reworded items, 5. GPT-4, original items, 6. GPT-4, reworded items, 7. GPT-4 mini, original items, 8. GPT-4 mini, reworded items, 9. GPT-5, original items, 10. GPT-5, reworded items, 11. GPT o1, original items, 12. GPT o1, reworded items, 13. GPT o3, original items, 14. GPT o3, reworded items, 15. Llama, original items, 16. Llama, reworded items, 17. CENTAUR, original items, 18. CENTAUR, reworded items. p-values below the conventional 0.05 level are highlighted in bold for emphasis.}
\end{table}

\subsection*{Analysis of Rating Shifts}

To assess the strength of shifts in ratings in response to wording changes, we calculated the mean absolute difference in ratings between original and reworded items for each scenario, separately for LLMs and human raters. Across items, GPT-4 exhibited a mean absolute rating shift of 1.39 ($SD = 1.33$). GPT-4-mini showed a mean shift of 0.43 ($SD = 0.56$), comparable to the mean shift of 0.75 ($SD = 1.47$) shown by GPT-3.5. Newer models exhibited larger mean shifts: 2.16 ($SD = 1.72$) for GPT-5, 1.53 ($SD = 1.40$) for GPT-o1, and 1.84 ($SD = 1.44$) for GPT-o3. Llama-3.1 showed a mean shift of 1.25 ($SD = 1.47$), and CENTAUR a comparable shift to Llama-3.1 of 1.18 ($SD = 1.38$). Human ratings, by contrast, displayed the largest differences, with a mean shift of 2.20 ($SD = 1.08$). This supports, once more, our argument that LLMs do not react to reworded items the same way humans do.

\begin{figure}
\centering
\includegraphics[width=\textwidth]{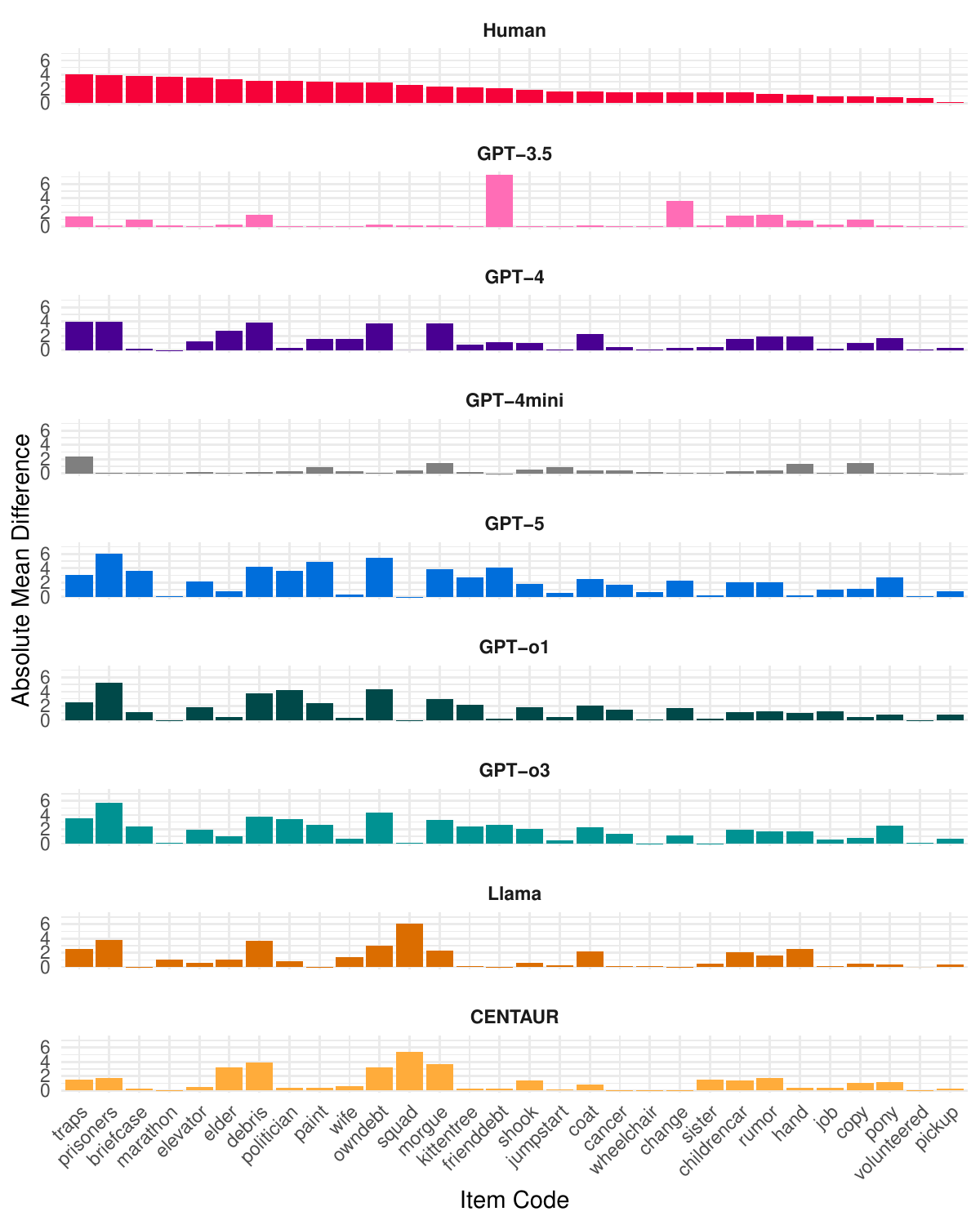}
\caption{Rewording-induced Change in Morality Ratings. Note: Bar plots show the absolute mean difference between ratings of original scenarios and their reworded counterparts (x-axis), plotted separately for (from top to bottom) human ratings, GPT-3.5, GPT-4, GPT-4-mini, GPT-5, GPT-o1, GPT-o3, Llama and CENTAUR. To ensure comparability, items are sorted within each panel to reflect the decreasing magnitude of change for human ratings, GPT-3.5, GPT-4, Llama and CENTAUR. To ensure comparability, items are sorted within each panel to reflect the decreasing magnitude of change for human ratings.}
\label{fig:abs_mean_diff}
\end{figure}

Comparing the shifts per scenario in detail (see Figure~\ref{fig:abs_mean_diff}) also reveals inconsistencies between models: GPT-4 and GPT-3.5 both tend not to react much to semantic rewording but they react to different scenarios. GPT-3.5 tends to be slightly more in line with human shifts compared to GPT-4. Intriguingly, GPT-5, GPT-o1 and GPT-o3 show qualitatively comparable patterns per item. Llama and CENTAUR react more to changes, but quite differently than humans. Changes between Llama and CENTAUR appear consistent, which may be explained by the fact that CENTAUR is a fine-tuned version of Llama, although the present analysis cannot establish the source of that similarity. A post-hoc, exploratory analysis that grouped items by rewording category (Supplementary Material, Section~\ref{sec:supp:s2}) revealed that, descriptively, mean shifts vary across categories and models, but category-level effects were not statistically robust, suggesting largely item-specific effects.

%% file: discussion.tex
Our findings provide an empirical illustration of our argument that LLMs may react to novel stimuli differently compared to human participants across eight LLMs (GPT-3.5, GPT-4, GPT-4.0-mini, GPT-5, GPT-o1, GPT-o3, Llama-3.1 70b and CENTAUR). Our key result is that the similarity between average human responses and average LLM responses reduces significantly when items are reworded.

Across LLMs, the magnitude of the divergence from human responses varied. Older LLMs tended to simply ignore changes in wording, whereas modern LLMs displayed different, sometimes idiosyncratic, patterns. To provide some illustrative examples: Humans regarded it as much less moral to take a shortcut during a marathon in order to win compared to running on the course during the race, whereas LLMs largely viewed them as equally moral. Similarly, humans gave a clearly negative rating to a politician using federal tax money for building an extension on their own home, but a neutral rating to a politician using federal tax money to extend their city. Interestingly, GPT-3.5, GPT-4, GPT-4-mini, Llama, CENTAUR generally judged both actions as unethical, whereas newer models of the GPT-family (GPT-5, GPT-o1, GPT-o3) distinguished them, rating the private-home extension as highly unethical, but treated the city extension as mildly-to-moderately ethical. These examples highlight how LLMs can either overlook, or sometimes unrealistically amplify meaningful ethical distinctions that humans make.

Yet, some models---notably GPT-5 and GPTo3---did react to rewording and showed closer alignment with human ratings in many comparisons compared to older models. It is tempting to take this as evidence that newer LLMs are genuinely more human-like. There are, however, reasons for caution. First, our core result persisted: the similarity between LLM ratings and human ratings was significantly reduced for reworded items. Second, these models still failed to capture the variance of human responses. Figure~\ref{fig:mean_ratings} shows that humans tend to disagree on morality ratings, while LLMs tend to provide the same rating across repeats. This difference in variance is likely to be even more pronounced for more diverse samples of human participants (e.g.\ shaving a local elder may be considered immoral by human participants from different cultural backgrounds). Third, we only observed the improved alignment on these 30 example scenarios. It may well be that more recent fine-tuning efforts included examples of such rewording (such as the preprint of this paper) in the training data, thus enhancing robustness in these specific cases. However, our argument gives reason to suspect that counterexamples can still be found, even if the specific reworded items in our experiment should no longer produce the same divergence in future models. Hence, we advise researchers to validate the similarity between LLM responses and human responses for their specific experiment if using LLMs at all.

\subsection{Limitations}

Our study is not without limitations. First, because human ratings were collected in a between-subjects design, these observed differences necessarily reflect both sensitivity to rewording and additional between-participant variability; accordingly, our analyses focus on group-level patterns of change rather than within-individual effects. Additional analyses indicated that human responses were more variable in the reworded condition (see Section~\ref{sec:supp:s3} of the Supplementary Material), consistent with increased ambiguity. However, when restricting comparisons to items with similar mean ratings, this difference was no longer reliable, suggesting that it was partly driven by distributional differences. Crucially, these effects do not fully account for the observed differences in how humans and LLMs respond to rewording at the item level. Furthermore, heterogeneity in human responses is a substantive feature of human psychological data rather than mere noise. This point adds to the criticism of LLMs as simulators for human participants: Prior work has shown that LLMs systematically fail to reproduce this variance in responses, instead tending toward uniform or overly stable judgments \citep{harding_ai_2024, kwok_evaluating_2024, rime_2025, wang_what_2025}.

A second limitation of our study concerns the rewording manipulation itself: while we controlled item length and aimed to preserve overall structure, the perturbations are not ``uniform'' in the sense of forming a single interchangeable class.\footnote{To further examine whether different types of rewording account for variability in rating shifts, we conducted a post hoc categorization of items into five perturbation classes (Fantasy, Ambiguity, Cultural nuance, Neutral, Unexpected), based on independent coding by three researchers (majority label; disagreements resolved by discussion). Descriptively, some categories appeared to induce larger shifts; however, non-parametric tests did not indicate reliable differences across categories (all $p > .12$). Full details and figures are provided in Section~\ref{sec:supp:s2} of the Supplementary Material.} Some rewording introduces implausible or fantastical elements (e.g., ``leaving earth''), others increase referential ambiguity (e.g., unclear pronouns), and others may shift culture-specific connotations (e.g., judgments about killing rats or being late for a wedding). Accordingly, we do not interpret our results as isolating the causal impact of any single perturbation category. Rather, we perform a heterogeneous stress test of robustness to various types of rewording, and Figure~\ref{fig:abs_mean_diff} illustrates that item-level shifts vary across stimuli and LLMs.

Third, we acknowledge that we were unable to identify a satisfying explanation for the difference between the responses of human participants and more modern LLMs such as GPT-5 and GPT-o3. Whereas older LLMs seemed to mostly ignore rewording, apparently over-generalizing from the training data in a simple fashion, the responses of GPT-5 and GPT-o3 did change---but in idiosyncratic ways that differed from human reactions. This finding relates to the broader explainability problem of LLMs: it is challenging to explain why an LLM responded the way it did \citep{zhao_explainability_2024}.

Overall, we acknowledge that the observed results may be due to a mixture of effects, including differences in the task (original versus reworded items), different perturbation categories, and differences between humans and LLMs in processing rewording. Our data cannot fully disambiguate these effects. Still, irrespective of the precise underlying cause, the reduction of similarity between LLM and human answers calls for caution when using LLMs in research.

\subsection{Conclusion}

Taken together, our empirical results and theoretical argument lead us to recommend that psychologists should refrain from using LLMs as replacement for human participants---in line with ample prior work (e.g., \citet{van_rooij_combining_2025, lin_six_2025}). If human responses are augmented with LLM-generated data, LLM outputs should be carefully calibrated against human responses (refer, e.g., to \citet{egami_using_2023}). Still, consistent with more optimistic perspectives such as \citet{dillion_can_2023} and \citet{binz_foundation_2025}, we acknowledge that LLMs may be supportive in other parts of the research process, such as pilot testing when pilot testing with human participants is not feasible. For research questions that aim to characterize human judgments, experiences, or behavior, however, human participants cannot be replaced.

%% file: supplement.tex
% Supplementary Material for:
% Schroeder, Morgenroth, Kuhl, Vaquet, and Paaßen (2026)
% "Exploring the Limits of Large Language Models as a Reflection of Human
%  Cognition: An Illustration in the Context of Moral Judgement"

\section{Supplementary Material}\label{sec:supplement}

In this Supplement, we provide additional robustness checks and exploratory analyses that complement the main manuscript. Section~\ref{sec:supp:s1} reports an auxiliary regression-based comparison of pooled versus group-specific fits (humans vs.\ each LLM) to assess whether allowing separate regression relationships improves fit beyond chance under label permutation. Section~\ref{sec:supp:s2} addresses concerns about non-uniform stimulus perturbations by introducing a post-hoc coding scheme for rewording categories and examining whether rating shifts systematically differ across these categories. We present descriptive patterns and non-parametric tests. Section~\ref{sec:supp:s3} evaluates an alternative explanation centered on distributional differences that may point to more foundational task differences between conditions. Specifically, we tested for variance heterogeneity in human responses and conducting mean-yoked matching analyses to assess the extent to which differences in item-mean distributions contribute to observed variability.

\subsection{Comparison of model fits between a pooled linear regression (combining LLM and human data) and a group-specific models}\label{sec:supp:s1}

\subsubsection{Motivation and Method}\label{sec:supp:s1:method}

As a post hoc test for cases where we did not find evidence that humans change their ratings more than LLMs do, we evaluated whether the change from original to reworded items was the same for LLMs and human participants by comparing model fits between a pooled linear regression (combining LLM and human data) and a group-specific model with separate slopes for each group.

\begin{figure}[htbp]
\centering
\includegraphics[width=\textwidth]{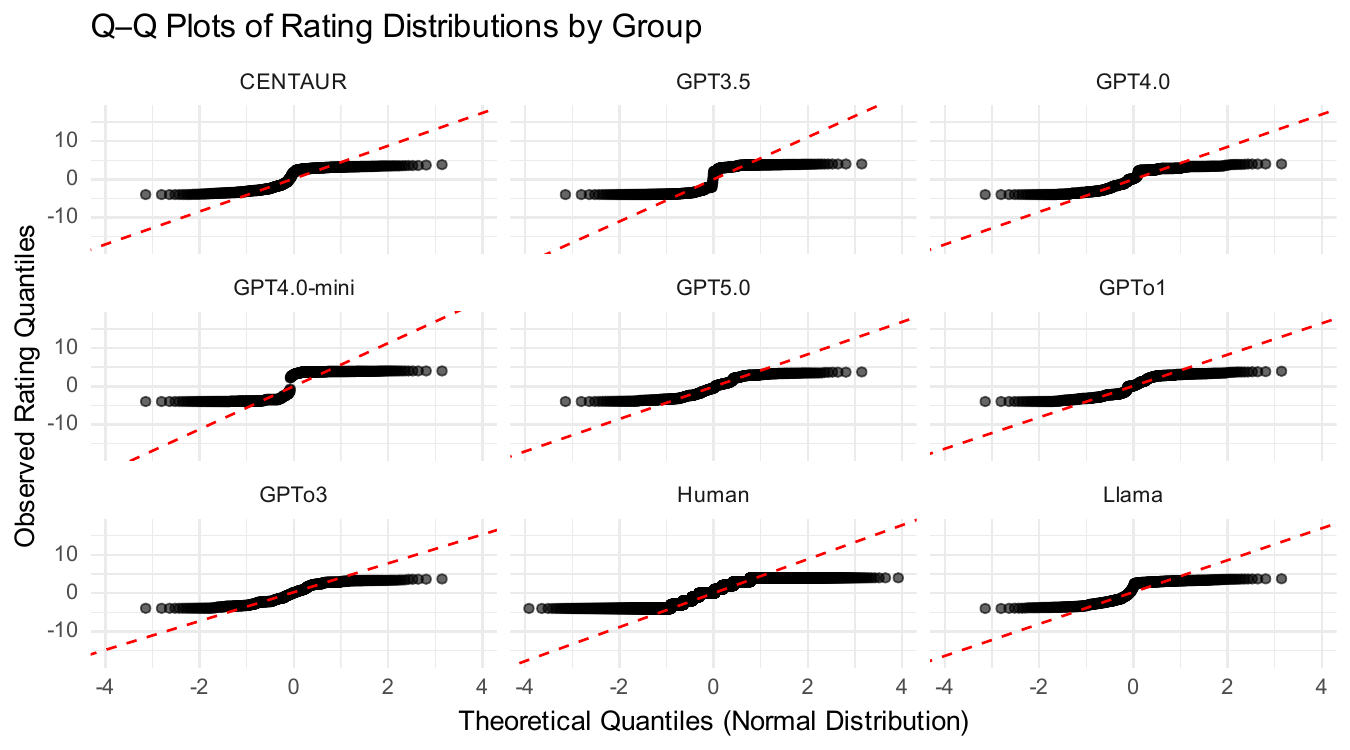}
\caption{Quantile-quantile plots of moral ratings by group/model. Observed rating quantiles per group are plotted against normal-theory quantiles; deviations from the diagonal indicate departures from normality.}
\label{fig:supp:s1:qq}
\end{figure}

Because normality assumptions were not met (Figure~\ref{fig:supp:s1:qq}), we assessed differences between pooled and group-specific regression fits using a permutation-based Chow test in which group labels (Human vs.\ LLM) were randomly permuted to generate an empirical null distribution of the Chow statistic \citep{chow_tests_1960}, estimating an empirical $p$-value for the fit improvement when moving from the pooled to the group-specific model; we report both the empirical $p$-values and the corresponding change in explained variance ($\Delta R^2$). If LLMs simulate human judgement, the pooled model should suffice; if not, separate models will fit significantly better, thus indicating systematic divergence under slight changes in item wording.

\subsubsection{Results}\label{sec:supp:s1:results}

\begin{figure}
\centering
\includegraphics[width=\textwidth]{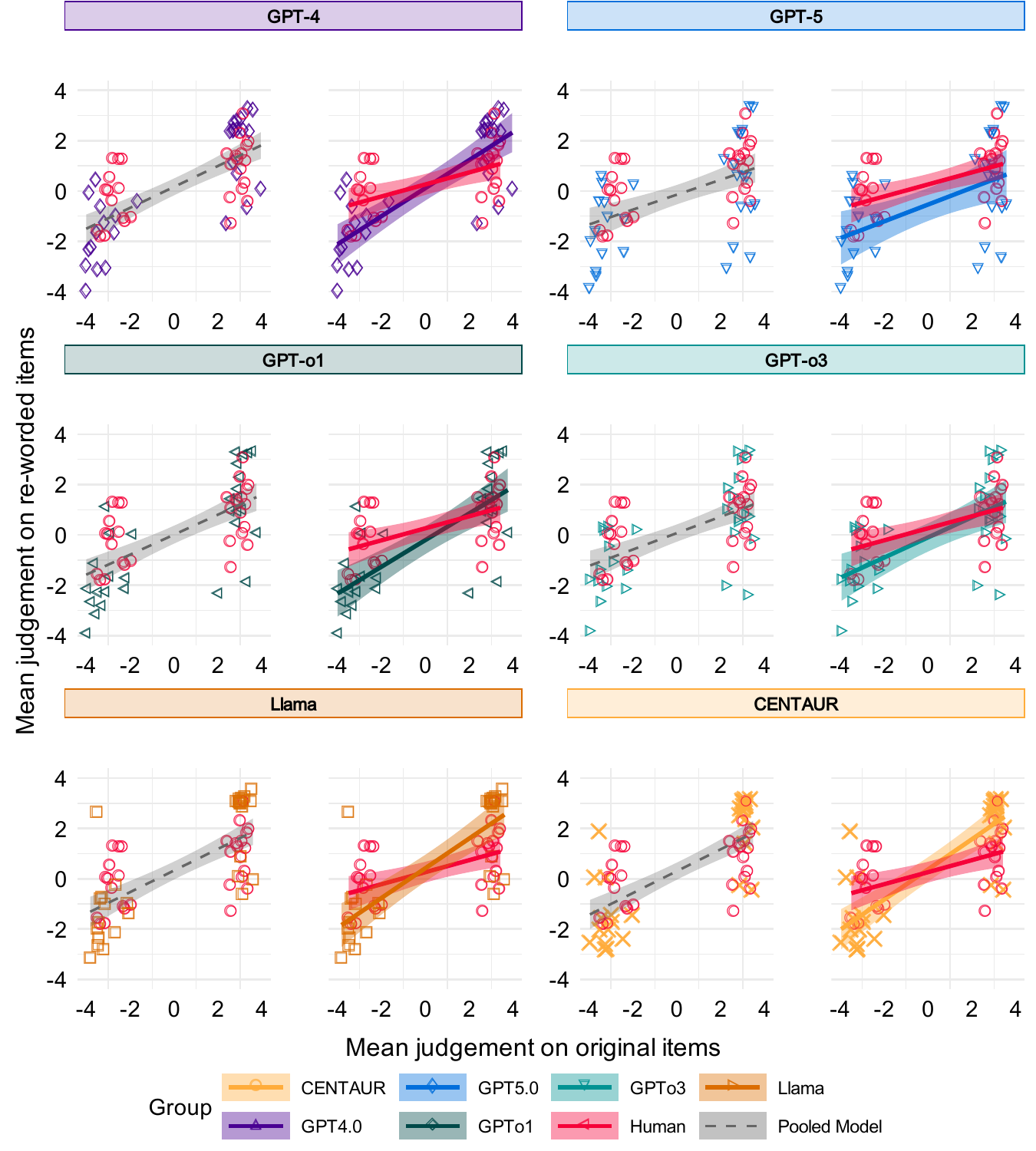}
\caption{Scatter Plot of Mean Morality Ratings for Each Scenario in the Reworded Condition (Y-Axis) Against its Rating to the Original Wording (X-Axis). \emph{Note.} Crosses mark CENTAUR ratings, diamonds mark GPT-4 ratings, downward triangles mark GPT-5 ratings, left-pointing triangles mark GPT-o1 ratings, right-pointing triangles mark GPT-o3 ratings, circles mark human ratings, and squares mark Llama ratings. The same points are plotted in each subplot per model, showing the pooled fit through all points (grey), contrasted with group-specific regressions. Shaded areas around the regression lines indicate the 95\% confidence interval.}
\label{fig:supp:s1:scatter}
\end{figure}

We compared model fits between single pooled regressions (combining human and model data) with group-specific regressions for the six top-performing LLMs in terms of human-likeness (Figure~\ref{fig:supp:s1:scatter}), GPT-4, GPT-5, GPT-o1, GPT-o3, Llama and CENTAUR using a permutation-based variant of Chow test that does not rely on parametric assumptions (Table~\ref{tab:supp:s1:chow}). For Llama and CENTAUR, fitting separate lines for human and LLM ratings led to a significantly better fit than the pooled model ($F = 5.47$, $p = .006$, and $F = 6.36$, $p = .004$). Likewise, for GPT-4 ($F = 4.78$, $p = .013$) and GPT-o1 ($F = 4.05$, $p = .023$), the separate-group regressions outperformed the pooled regression. In these cases, wording changes influenced human and LLM ratings in distinct ways, in line with our argument. Intriguingly, however, our analysis did not show evidence that separate-group regressions fit the data better than a pooled model for GPT-5 ($F = 2.45$, $p = .095$) and GPT-o3 ($F = 1.46$, $p = .240$). As shown in Table~\ref{tab:supp:s1:r2}, allowing group-specific regressions substantially improved model fit for GPT-4 ($\Delta$ adjusted $R^2 = .38$), Llama ($\Delta$ adjusted $R^2 = .16$), and CENTAUR ($\Delta$ adjusted $R^2 = .19$), yielded moderate improvements for GPT-o1 ($\Delta$ adjusted $R^2 = .12$) and GPT-o3 ($\Delta$ adjusted $R^2 = .04$), and provided no improvement for GPT-5 ($\Delta$ adjusted $R^2 = .00$), indicating that only GPT-5 is adequately captured by a single pooled model.

\begin{table}
\centering
\caption{Results of the permutation-based Chow-type tests comparing pooled vs.\ group-specific regressions (humans + LLM vs.\ humans and LLMs separately), for the top-performing models in human-likeness.}
\label{tab:supp:s1:chow}
\begin{tabular}{@{}lcc@{}}
\toprule
Model & Test statistic ($F$) & Empirical $p$-value \\
\midrule
GPT-4     & 4.78 & \textbf{.013} \\
GPT-o1    & 4.05 & \textbf{.023} \\
GPT-5     & 2.45 & .095 \\
GPT-o3    & 1.46 & .240 \\
Llama     & 5.47 & \textbf{.007} \\
CENTAUR   & 6.36 & \textbf{.004} \\
\bottomrule
\end{tabular}
\medskip

{\small\emph{Note.} Empirical $p$-values are obtained from a permutation-based Chow-type structural-break test (group labels permuted) comparing a pooled regression (humans + LLM) to a group-specific regression allowing separate slopes by group. $p$-values below the conventional 0.05 level are highlighted in bold for emphasis and suggest that fitting separate lines for human and LLM ratings led to a significantly better fit than a pooled model.}
\end{table}

\begin{table}
\centering
\caption{Adjusted $R^2$ for pooled vs.\ group-specific regressions.}
\label{tab:supp:s1:r2}
\begin{tabular}{@{}lcccc@{}}
\toprule
LLM & Pooled: Human + LLM & Human only & LLM only & LLM only $-$ Pooled \\
\midrule
GPT-4   & .27 & .27 & .65 & +.38 \\
GPT-5   & .27 & .27 & .27 & .00 \\
GPT o1  & .43 & .27 & .55 & +.12 \\
GPT o3  & .34 & .27 & .39 & +.04 \\
Llama   & .47 & .27 & .63 & +.16 \\
CENTAUR & .50 & .27 & .69 & +.19 \\
\bottomrule
\end{tabular}
\end{table}

\subsubsection{Interpretation}\label{sec:supp:s1:interpretation}

Across the top-performing models, this test identifies significant structural breaks for GPT-4, GPT-o1, Llama, and CENTAUR, indicating that human and model ratings are better described by separate regression relationships under rewording. In contrast, GPT-5 and GPT-o3 do not show reliable evidence for a structural break, aligning with their comparatively closer human-like sensitivity by this particular diagnostic. Importantly, these results are best interpreted as supplementary to $\Delta R^2$: they speak to statistical detectability under rewording rather than the magnitude of the divergence, which is better captured by the variance-explained metrics in the main manuscript.

\subsection{Category-wise Analysis of Rewordings}\label{sec:supp:s2}

\subsubsection{Motivation and Method}\label{sec:supp:s2:method}

To follow up on reviewer feedback regarding heterogeneity in perturbation categories, we conducted an exploratory item-category analysis to examine whether different categories of rewording systematically account for variation in rating shifts between original and reworded items.

Categories were derived iteratively from the set of reworded items by identifying recurring patterns and consolidating them into a unified scheme comprising five categories: Fantasy, Ambiguity, Cultural nuance, Neutral, and Unexpected (Table~\ref{tab:supp:s2:categories}).

Three researchers independently coded all items using this scheme. The majority label was used as the final category assignment; cases of complete disagreement were resolved through discussion. The resulting distribution of items was unbalanced (C = 4, A = 4, F = 5, N = 11, U = 6), reflecting the post hoc nature of this categorization. Table~\ref{tab:supp:s2:categories} gives definitions and illustrative examples for the item categories.

\begin{table}
\centering
\caption{Definitions and illustrative examples for the categories used to characterize rewordings.}
\label{tab:supp:s2:categories}
\begin{tabular}{@{}p{2.5cm}p{6cm}p{6.2cm}@{}}
\toprule
Category & Definition & Example \\
\midrule
Neutral (N) &
Rewording preserves a realistic scenario without obvious
plausibility shifts, ambiguity, or culture-specific assumptions. &
marathon: ``Person X ran on the course during the marathon in
order to win.'' \\
\addlinespace
Fantasy (F) &
Rewording introduces implausible or fantastical elements,
changing the scenario's realism. &
wife: ``Person X blocked his wife from leaving earth or
interacting with aliens.'' \\
\addlinespace
Ambiguity (A) &
Rewording introduces uncertainty of how the scenario is to be
interpreted (e.g., unclear referent, underspecified action, or
multiple plausible readings). &
paint: ``Person X offered to help an elderly neighbor paint
his dog.'' \\
\addlinespace
Cultural Nuance (C) &
Rewording shifts the scenario into a context where moral
evaluation plausibly depends on cultural norms or background
knowledge. &
elder: ``Person X cut the beard off of a local elder to shave
him.'' \\
\addlinespace
Unexpected (U) &
Rewording introduces an unusual premise that is plausible but
atypical, thus introducing surprising contextual details rather
than a clean semantic paraphrase. &
cancer: ``Person X shaved their leg when they found out their
partner had cancer and required radiation therapy.'' \\
\bottomrule
\end{tabular}
\end{table}

For each item and group, we computed the difference between ratings for reworded and original items. We then examined how these shifts varied across categories and groups. Finally, to assess whether perturbation categories systematically influenced shift magnitudes, we conducted Kruskal--Wallis tests within each group, comparing absolute shift values across categories.

\subsubsection{Results}\label{sec:supp:s2:results}

\begin{figure}
\centering
\includegraphics[width=\textwidth]{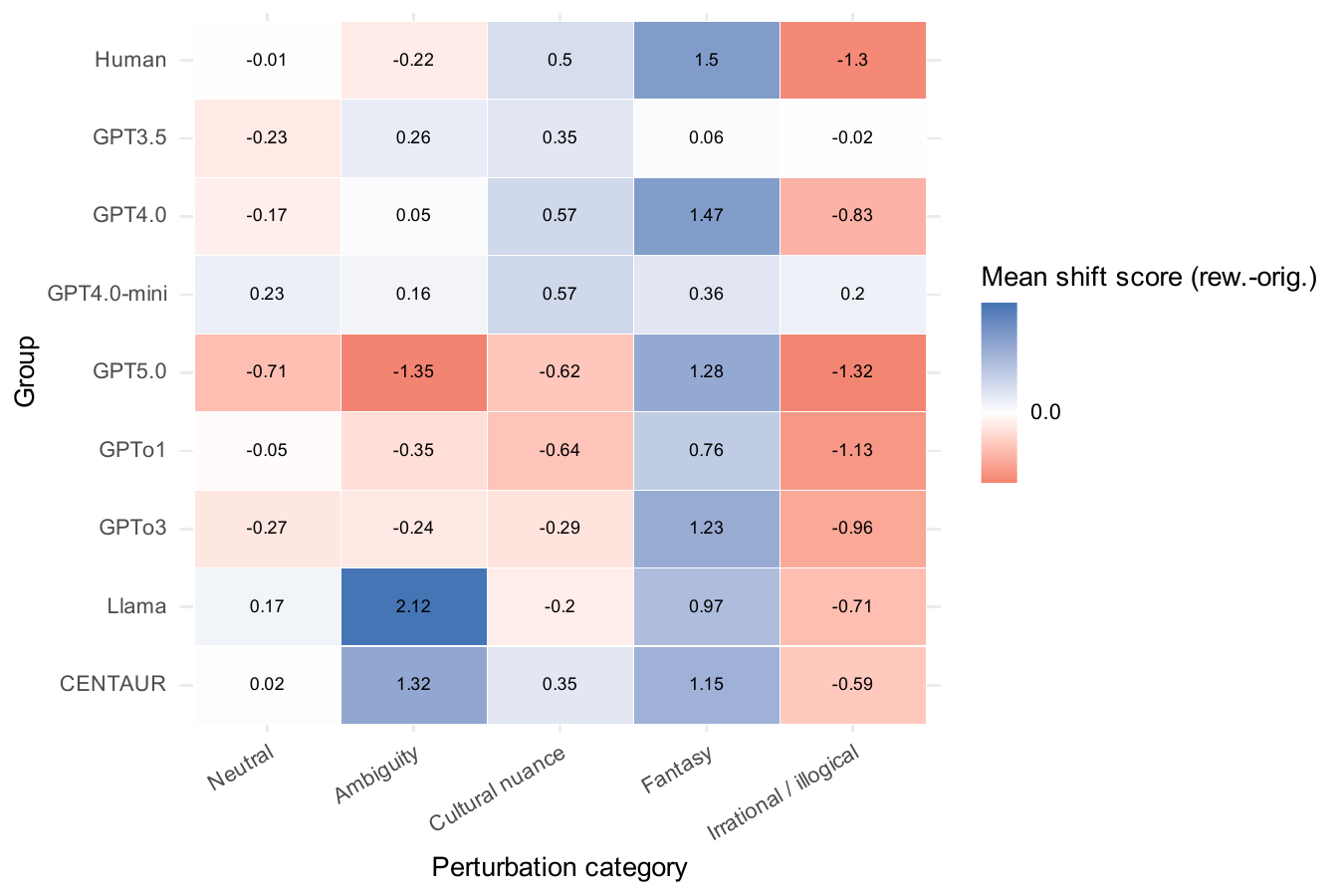}
\caption{Mean shift in morality scores per group per category (reworded $-$ original). Blue tiles suggest an average increase in morality rating after rewording; red hues suggest a decrease in morality rating after the modification.}
\label{fig:supp:s2:heatmap}
\end{figure}

Figure~\ref{fig:supp:s2:heatmap} shows the mean shift between reworded and original items for each category and group.

Descriptively, the magnitude and direction of mean shifts differ systematically across categories and models. In particular, some categories tend to induce larger shifts across groups that show a certain degree of consistency: for fantasy rewordings, humans and LLMs showed a tendency to react with higher morality judgements, while unexpected rewordings led to lower morality judgements for unexpected items. So, overall, these descriptive patterns indeed differ visually between humans and LLMs, indicating that---qualitatively---model insensitivity is not uniform across categories.

However, non-parametric tests did not reveal statistically significant differences in shift magnitudes across categories for any group (all $p > .12$; all Holm-corrected $p = 1$). Thus, while item-level variability is evident, it does not organize into robust, category-level effects.

\begin{figure}
\centering
\includegraphics[width=\textwidth]{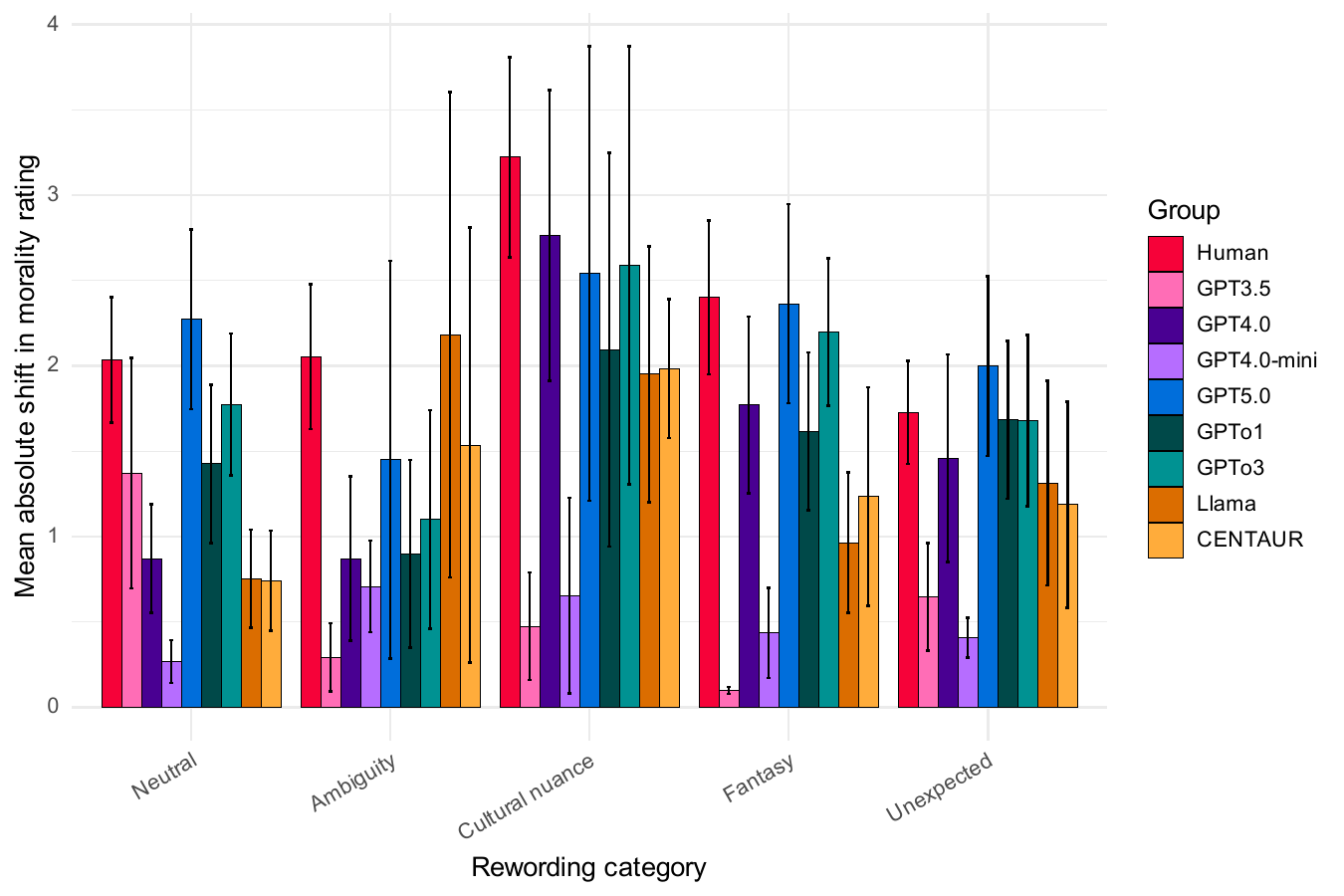}
\caption{Mean shift in absolute morality scores per group per category (reworded $-$ original). Bars show the mean absolute change in morality ratings between original and reworded versions, computed within each rewording category (Neutral, Ambiguity, Cultural nuance, Fantasy, Unexpected) for humans and each LLM. Error bars indicate the standard error of the mean.}
\label{fig:supp:s2:barchart}
\end{figure}

\subsubsection{Discussion and Conclusion}\label{sec:supp:s2:discussion}

These findings suggest that the impact of rewording is not easily reducible to a small set of categories. Instead, sensitivity to rewording appears to be largely item-specific, consistent with the main interpretation of the manuscript.

At the same time, the descriptive patterns indicate that different categories of perturbations may still influence human and LLM judgments in partially distinct ways: For neutral, ambiguous, and culturally nuanced rewordings, changes were inconsistent between humans and LLMs and between different LLM models. For fantasy or unexpected rewordings, however, shifts of humans and LLMs where at least directionally similar. Given the limited and unbalanced number of items per category, these results should be interpreted cautiously.

Taken together, this exploratory analysis provides a first step toward understanding how different perturbation categories affect model--human alignment. A more definitive test would require a design with pre-specified, balanced perturbation categories and multiple controlled rewordings per scenario.

\subsection{Variability and Distributional Differences Across Conditions}\label{sec:supp:s3}

\subsubsection{Motivation and Method}\label{sec:supp:s3:method}

To address the possibility that differences between original and reworded conditions reflect distributional or task-related variability rather than sensitivity to semantic changes, we conducted additional analyses examining the variability of human responses across conditions.

We first assessed whether the variability of human ratings differed between original and reworded conditions using a Fligner--Killeen test of homogeneity of variances. This non-parametric test is robust to non-normality and therefore appropriate for the present data.

To further examine whether this difference is driven by the distribution of item means, we implemented a mean-yoked comparison. Specifically, we performed optimal one-to-one matching between original and reworded items based on item mean ratings, minimizing absolute mean differences across pairs. Matching was restricted using calipers (maximum allowed mean difference) of 0.25, 0.5, and 1.0 scale units. Items without suitable matches were excluded.

Variability (standard deviations) was then compared across matched pairs using paired Wilcoxon signed-rank tests.

\subsubsection{Results}\label{sec:supp:s3:results}

\begin{table}[b]
\centering
\caption{Paired Wilcoxon signed-rank tests comparing variability (standard deviation) of human ratings between original and reworded conditions on mean-yoked item pairs.}
\label{tab:supp:s3:calipers}
\begin{tabular}{@{}lcc@{}}
\toprule
Caliper & Matched pairs ($N$) & $p$-value \\
\midrule
0.25 & 3  & .181 \\
0.50 & 5  & .178 \\
1.00 & 11 & .056 \\
\bottomrule
\end{tabular}
\end{table}

The Fligner--Killeen test indicated significantly greater variability in the reworded condition across all items for human participants ($\chi^2 = 861.43$, $p < .001$). This result suggests that reworded items elicit more variable human responses, consistent with increased ambiguity or task complexity.

The comparison of mean-yoked items with optimal matching showed that across all calipers, differences in variability were not statistically significant, although a trend toward higher variability in the reworded condition was observed for more lenient matching criteria.

\subsubsection{Interpretation}\label{sec:supp:s3:interpretation}

These results indicate that the increased variability observed in the reworded condition is partly attributable to differences in the distribution of item means. When controlling for mean levels, the variance difference is no longer robustly detectable, although these analyses are based on substantially reduced sample sizes and thus limited statistical power.

Importantly, these findings do not account for the main results of the study, which are based on item-level comparisons of rating changes between original and reworded versions of the same scenarios. These analyses reveal systematic differences in how humans and LLMs respond to rewording, beyond what can be explained by distributional differences alone.

\subsection{Prompt}
\label{sec::supp:s4}

\begin{verbatim}
On the next page, you will be presented with a range of different behaviors.
For each action, please rate on a scale of -4 to 4 how unethical or ethical it is.
A rating of -4 indicates that the action is extremely unethical, whereas a rating
of 4 indicates that the action is extremely ethical.
Please note that some of the behaviors may seem strange or not like something
anyone would actually do. There might be surprising words or behaviors that
seem bizarre. These are not errors.
Please still rate how ethical or unethical the action is.
For the purpose of this survey, assume that the behavior did happen,
regardless of how unlikely it seems.

Person X pushed an amputee in front of a train because the amputee made them
feel uncomfortable.
Rating: -3.78

Person X ignored a new person in the office because they thought the new person
was Jewish.
Rating: -3.07

Person X smeared dog poo on their colleague’s chair and laughed uncontrollably
when their colleague sat on it.
Rating: -2.54

Person X heckled a stranger and made cruel remarks just because the stranger was
overweight.
Rating: -2.83

Person X pushed in front of another patron in the line at a theater.
Rating: -1.80

Person X got a friend to fix their gutters and promised guitar lessons in return,
but never honored the promise.
Rating: -1.96

Person X forgot they had to attend their niece’s dance concert that evening.
Rating: -0.44

Person X never learned how to ride a bicycle.
Rating: -0.01

Person X has memorized three of Shakespeare’s plays.
Rating: 0.28

Person X locked their keys in the house and had to call a locksmith.
Rating: 0.19

Person X learned an impressive dance routine in preparation for a friend’s wedding.
Rating: 1.02

Person X went skydiving despite their intense fear of heights because it was
their sister’s wish to do it together.
Rating: 1.52

Person X repaid a loan of $100 that their friend had lent them, even though the
friend did not remember it.
Rating: 2.98

Person X saw someone across the road drop a stack of papers, so they crossed the
road to help.
Rating: 2.61

Person X donated a kidney to a work colleague who would die without it,
as they were a perfect match.
Rating: 3.51

Person X pulled over on a busy highway on a rainy day to help a stranger
change his flat tire.
Rating: 3.17

<TEST SENTENCE>
\end{verbatim}